\documentclass[conference]{IEEEtran}
\IEEEoverridecommandlockouts
\usepackage{cite}
\usepackage{amsmath,amssymb,amsfonts}
\usepackage{algorithmic}
\usepackage{graphicx}
\usepackage{textcomp}
\usepackage{xcolor}
\usepackage{float}
\usepackage{booktabs}
\usepackage{amssymb}
\usepackage[table]{xcolor}
\usepackage{booktabs}
\usepackage{float}
\usepackage{url}

\definecolor{vlmcolor}{HTML}{0072B2}
\definecolor{reactcolor}{HTML}{D55E00}
\definecolor{trucolor}{HTML}{CC79A7}

\def\BibTeX{{\rm B\kern-.05em{\sc i\kern-.025em b}\kern-.08em
    T\kern-.1667em\lower.7ex\hbox{E}\kern-.125emX}}
\begin{document}

\title{Evaluating Long-Horizon Robot Planning under Uncertainty\\
}

\author{Harsh Muriki$^{*1}$, Gayathri Aravindan$^{1}$, Yuchen Zheng$^{1}$, Sonia Chernova$^{1}$
\thanks{*Corresponding author}
\thanks{$^{1}$Georgia Institute of Technology, Atlanta, GA, USA
{\tt\small
\{vmuriki3, garavindan3, fzheng, chernova\}@gatech.edu}.
}}

\maketitle

\begin{abstract}


Household robots must successfully navigate imperfect circumstances, both in their internal models and the external world.
We introduce a benchmark of seven long-horizon household tasks in Habitat that vary initial robot memory and environment configurations.
We evaluate adaptations of ReAct, VLM-TAMP, and Tru-POMDP within a shared execution framework across 90 tasks--environment--memory combinations, totaling 1,350 episodes.
Our evaluation measures task success, partial completion, and action efficiency, along with rule-based analysis of execution traces.
ReAct achieves the highest overall task success (45.3\%) and completion (71.7\%), while having accurate initial memory does not consistently yield the highest performance for every planner. We identify recurring placement errors, precondition errors, and redundant actions through trace analysis. These findings highlight differences in how the evaluated planners use initial knowledge and respond to uncertainty. Project page: \url{https://harshmuriki.github.io/leap-planning-uncertainty/}



\end{abstract}



\section{Introduction}

In real homes, robots rarely hold accurate beliefs about their surroundings. An object remembered on a countertop may have been moved into a cabinet, replaced with a similar item, or removed entirely. Each situation demands a different response: searching for the object, selecting a substitute, revising the plan, or recognizing that the goal is infeasible. Robust planning therefore requires robots to gather information, update their internal world representations, and adapt their plans as new observations become available~\cite{kaelbling2013integrated,garrett2020onlinereplanningbeliefspace}.

\begin{figure}[h]
    \centering
    \includegraphics[width=1\linewidth]{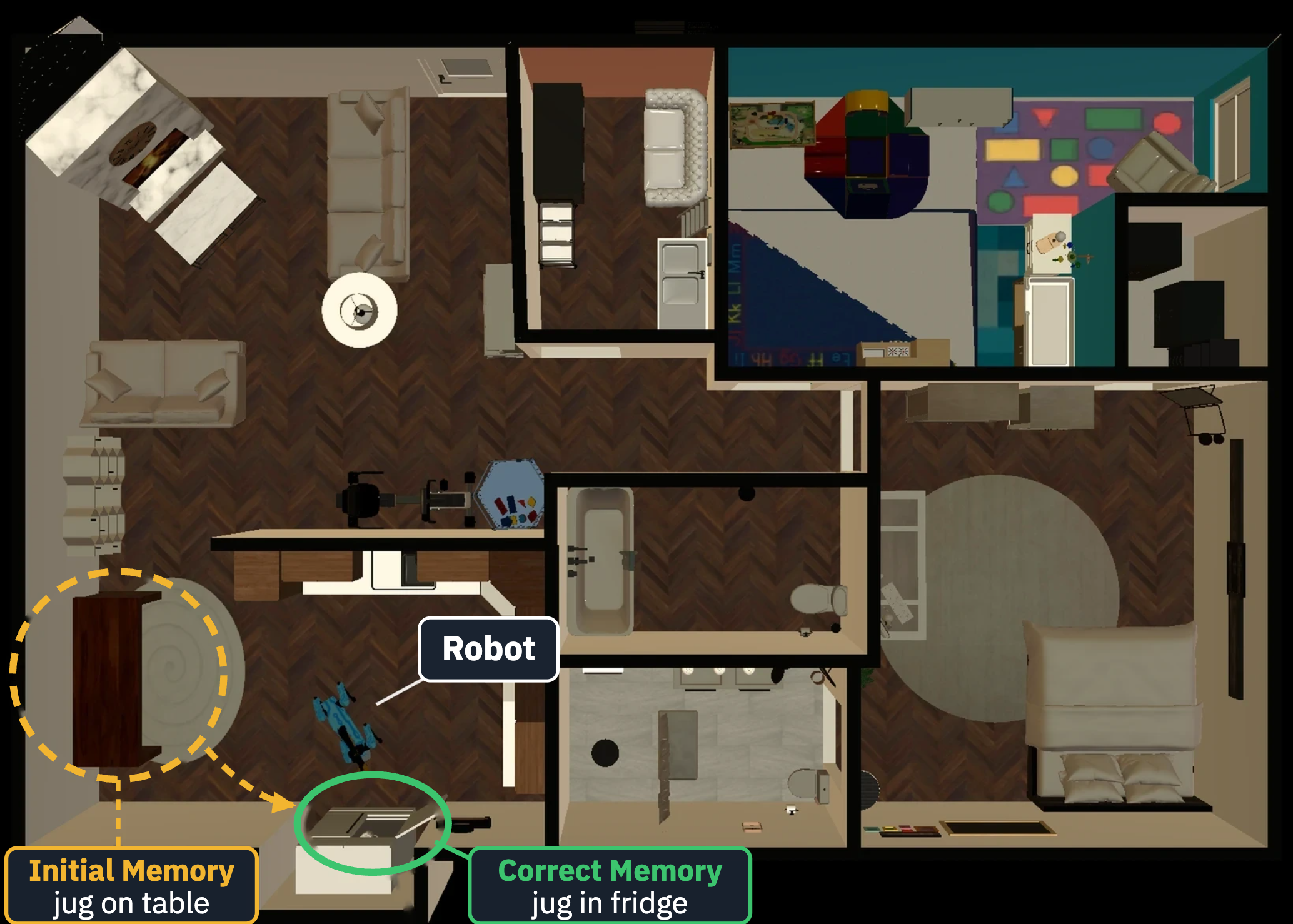}
    \caption{Task T1 under HID × Outdated: the robot remembers the jug on the dining table (gold), but it is hidden in the fridge (green).}
    \label{fig:diag-react}
\end{figure}




Task and motion planning (TAMP) connects discrete high-level task decisions with geometrically feasible robot motions~\cite{garrett2021integrated}; under partial observability, these decisions must also account for what the robot knows about the environment. Existing approaches address this challenge through different mechanisms.
ReAct~\cite{yao2023reactsynergizingreasoningacting} interleaves reasoning and action, while VLM-TAMP~\cite{vlmtamp-yang2024guiding} uses vision-language-model-generated subgoals to guide geometric planning. LLaMAR studies feedback-driven planning under partial observability~\cite{nayak2024LLaMAR}, and Tru-POMDP~\cite{tang2026trupomdptaskplanninguncertainty} explicitly models uncertainty through hypotheses and belief-space planning. Although these approaches offer different mechanisms for responding to uncertainty, their comparative strengths and limitations are difficult to characterize when evaluated on different tasks, environments, and execution frameworks. Benchmarks such as PARTNR~\cite{PARTNR} and Embodied Agent Interface~\cite{li2025embodiedagentinterfacebenchmarking} examine collaboration, embodied decision making, and execution errors. 
We investigate a complementary question: \textit{How do different planning approaches respond when initial robot memory and environment configurations are systematically varied within a shared execution framework?}

In this work, we investigate how different planning architectures respond to uncertainty in long-horizon household tasks. Specifically, we examine two complementary sources of uncertainty: \textit{initial memory uncertainty}, arising from incomplete or outdated knowledge of the environment, and \textit{environmental variation}, including unavailable objects, acceptable substitutions, hidden targets, and distractors. By systematically varying these conditions within a shared execution framework, we evaluate whether planners achieve their goals, how efficiently they adapt, and what causes execution failures.

We introduce a benchmark of seven long-horizon household tasks implemented in Habitat Sim~\cite{habitat19iccv,szot2021habitat,puig2023habitat3} using PARTNR's high-level robot skills. We adapt three planning approaches---ReAct, VLM-TAMP, and Tru-POMDP---to a common execution framework and evaluate them across 90 task--environment--memory combinations, with five repetitions per combination and planner, totaling 1,350 episodes. We measure task success, partial completion, and action efficiency relative to human reference executions, and analyze execution traces to identify recurring failure and inefficiency patterns.

Our results show that overall performance and sensitivity to uncertainty differ across the three planner adaptations. ReAct achieves the highest task success (45.3\%) and completion (71.7\%), outperforming approaches that incorporate explicit symbolic or belief-space planning. However, accurate initial memory does not consistently improve performance across planners, and all three approaches exhibit substantial execution inefficiencies. Trace analysis further reveals distinct limitations, including incorrect object placements, omitted actions, redundant exploration, and premature task termination. Together, these findings highlight that explicit uncertainty representations and replanning mechanisms alone do not guarantee robust or efficient execution.


In summary, our main contributions are:
\begin{itemize}
    \item A benchmark of seven long-horizon household tasks with controlled variations in initial robot memory and environmental conditions.
    \item A comparative evaluation of three planner adaptations, evaluated across 1,350 episodes using success, completion and action-efficiency metrics.
    \item A trace-based analysis identifying recurring failure and inefficiency patterns.
\end{itemize}

\section{Related Work}
\label{sec:related-work}

\subsection{Task Planning Under Uncertainty}

Robotic task planning under uncertainty requires reasoning with imperfect information and adapting plans as new evidence becomes available. Belief-space approaches maintain probabilistic state estimates and interleave planning, information gathering, and execution. CoCo-TAMP~\cite{coco-kim2026large} incorporates language-model commonsense knowledge into belief estimation, while Tru-POMDP~\cite{tang2026trupomdptaskplanninguncertainty} combines LLM-generated hypotheses with Bayesian belief updates and online POMDP planning. CLUE~\cite{clue-ravichandran2026activelyresolvingcontextualuncertainty} refines task-relevant hypotheses through closed-loop perception and interaction, while CURE~\cite{CURE-yin2025reliablellmbasedrobotplanning} estimates uncertainty in LLM-generated plans to inform whether tasks should be executed autonomously or completed with human assistance.

\subsection{Language- and Vision-Language-Based Task Planning}

Large language models (LLMs) provide semantic knowledge for translating natural-language instructions into robot action sequences, while vision-language approaches incorporate perceptual observations directly into the planning process. SayCan~\cite{saycan-ahn2022can} combines language-based skill selection with learned affordances, while ProgPrompt~\cite{singh2022progpromptgeneratingsituatedrobot} grounds plan generation through prompts specifying available actions, objects, and executable examples. The LLM planners in PARTNR-Planner~\cite{PARTNR} select perception and action tools using task instructions and an updated world graph.

Hybrid frameworks combine language-model reasoning with structured representations and classical planning. LLM+P~\cite{liu2023llmpempoweringlargelanguage} translates natural-language planning problems into PDDL and invokes a classical planner. VLM-TAMP~\cite{vlmtamp-yang2024guiding} uses VLM-generated subgoals to guide task and motion planning, querying the VLM again when a subgoal cannot be refined into a feasible plan. NESYRO~\cite{ahn2025reliablecodeaspoliciesneurosymbolicframework} combines language-model code generation with symbolic verification and interactive validation to establish task-relevant preconditions.

\section{Benchmark Tasks}
\label{sec:benchmark}

We introduce a benchmark comprising \textbf{seven} long-horizon household tasks implemented in Meta's Habitat simulator. The tasks span water fetching, package preparation, book collection, grocery organization, guest table preparation, meal delivery, and household cleanup. Each requires multi-step execution across multiple rooms and furniture locations, involving object transport, spatial placement, and object-state manipulation. All tasks draw from a shared library of twelve high-level robot skills, implemented using PARTNR-Planner~\cite{PARTNR}.
BASE task instructions are provided in Appendix~\ref{sec:appendix}.

To systematically evaluate planning under uncertainty, we vary two aspects of each episode: the robot's initial memory and the environment configuration.

\smallskip
\noindent \textbf{Initial robot memory.}
Each episode initializes the planner’s internal world representation under one of three memory conditions.
\textit{Accurate memory} provides correct task-relevant information about object locations, containment relations,
and states.
\textit{Incomplete memory} omits selected task-relevant information that must be acquired during execution.
\textit{Outdated memory} retains previously valid information that no longer matches the environment, requiring the planner to revise its representation using observations and action feedback.

\smallskip
\noindent \textbf{Environment configurations.}
We evaluate robustness to environmental variation across five scenario configurations, summarized in Table~\ref{tab:external-uncertainties}. These configurations vary object availability, accessibility, and the presence of distractors. In the baseline configuration (\textbf{BASE}), all target objects are present and directly accessible, with no distractors. The remaining configurations introduce distinct challenges or sources of task execution uncertainty: \textbf{SUB} replaces target objects with acceptable substitutes, requiring the robot to identify alternative ways to satisfy task goals; \textbf{ABS} removes target objects without providing suitable substitutes, requiring the robot to accept that a task goal is infeasible; \textbf{HID} hides target objects inside closed receptacles, requiring additional actions to locate and retrieve them; and \textbf{DIS} introduces irrelevant objects that may interfere with target identification and selection. Each configuration is instantiated for applicable tasks and evaluated under all three memory conditions.


\begin{table}[h]
\centering
\caption{Five environment configurations used to evaluate robustness.}
\label{tab:external-uncertainties}
\small
\setlength{\tabcolsep}{4pt}
\renewcommand{\arraystretch}{1.15}
\begin{tabular}{l|c|c|c||c|c}
    \hline
    Scenario & Target & Substitute & Absent & Hidden & Distractor \\
    \hline
    BASE & \checkmark &            &            &            &            \\
    \hline
    SUB  &            & \checkmark &            &            &            \\
    \hline
    ABS  &            &            & \checkmark &            &            \\
    \hline
    HID  & \checkmark &            &            & \checkmark &            \\
    \hline
    DIS  & \checkmark &            &            &            & \checkmark \\
    \hline
    \end{tabular}
\end{table}

In ABS scenarios, a desirable response is to search plausible locations, report the unavailable resource, and complete the remaining achievable task requirements rather than repeatedly attempting an infeasible action. Evaluation criteria requiring unavailable objects are excluded, and only the remaining achievable requirements are evaluated.


\smallskip

We evaluate each approach using task success, task completion, and execution efficiency. Let $i \in \{1,\ldots,N\}$ index evaluation episodes, $K_i$ denote the number of benchmark-defined evaluation criteria for episode $i$, and $z_{i,j}\in\{0,1\}$ indicate whether criterion $j$ is satisfied.

\smallskip
\noindent \textbf{Task success and completion.}
We measure task completion as the fraction of evaluation criteria satisfied:
\[
C_i = \frac{1}{K_i}\sum_{j=1}^{K_i}z_{i,j}.
\]
An episode is considered successful (task success, $S_i=1$) only if all criteria are satisfied; otherwise, $S_i=0$. We report the mean task success rate, $\frac{1}{N}\sum_i S_i$, and mean task completion rate, $\frac{1}{N}\sum_i C_i$. Both metrics range from 0 to 1 and weight all episodes equally. For episodes that time out, we use the latest recorded evaluation outcomes and execution counts.

\smallskip
\noindent \textbf{Execution efficiency.}
We adapt Success weighted by Path Length (SPL)~\cite{anderson2018evaluation} to measure efficiency in terms of high-level robot skill attempts rather than navigation distance. Specifically, we define action-based SPL ($\mathrm{SPL}_{\mathrm{act}}$), which combines binary task success with action efficiency relative to a human reference execution:
\[
\mathrm{SPL}_{\mathrm{act}}
=
\frac{1}{N}\sum_{i=1}^{N}
S_i\frac{A_i^{\mathrm{ref}}}
{\max(A_i,A_i^{\mathrm{ref}})},
\]
where $A_i$ is the number of skill attempts recorded during episode $i$, and $A_i^{\mathrm{ref}}$ is the corresponding reference action count from a human demonstration of the task variant.

Because binary success does not capture partial progress in long-horizon tasks, we additionally introduce task-completion-weighted SPL ($\mathrm{TC\text{-}SPL}_{\mathrm{act}}$), which replaces binary success with the fraction of evaluation criteria satisfied:
\[
\mathrm{TC\text{-}SPL}_{\mathrm{act}}
=
\frac{1}{N}\sum_{i=1}^{N}
C_i\frac{A_i^{\mathrm{ref}}}
{\max(A_i,A_i^{\mathrm{ref}})}.
\]
This metric rewards partial task completion while penalizing unnecessary skill attempts.

\smallskip
\noindent \textbf{Failure and inefficiency analysis.}
To characterize the sources of execution failures and inefficiencies, we apply rule-based analysis to episode traces, following prior work on embodied-agent error analysis~\cite{li2025embodiedagentinterfacebenchmarking, yang2025embodiedbenchcomprehensivebenchmarkingmultimodal}. We identify recurring patterns of failure and avoidable or redundant actions, and compare their prevalence across planners, initial memory conditions, and environment configurations.
Categories are not mutually exclusive; their definitions are provided in Appendix~\ref{sec:appendix}.


\section{Experimental Setup}
\label{sec:experimental-setup}


\subsection{Compared Methods}

We adapt three existing planning approaches to the Habitat/PARTNR framework, using a shared environment representation and library of executable robot skills to enable consistent comparison.

\smallskip
\noindent \textbf{ReAct adaptation.}
We extend the zero-shot ReAct planner~\cite{yao2023reactsynergizingreasoningacting} provided by PARTNR-Planner~\cite{PARTNR} to incorporate visual observations alongside task instructions, world graph descriptions, and action history, as shown in Figure~\ref{fig:diag-react}. At each decision step, the VLM generates a textual rationale and selects the next high-level robot skill. Execution feedback and updated observations are incorporated into subsequent decisions, enabling the planner to adapt its actions during execution. We refer to this multimodal adaptation as \textbf{ReAct}.

\begin{figure}[h]
    \centering
    \includegraphics[width=1\linewidth]{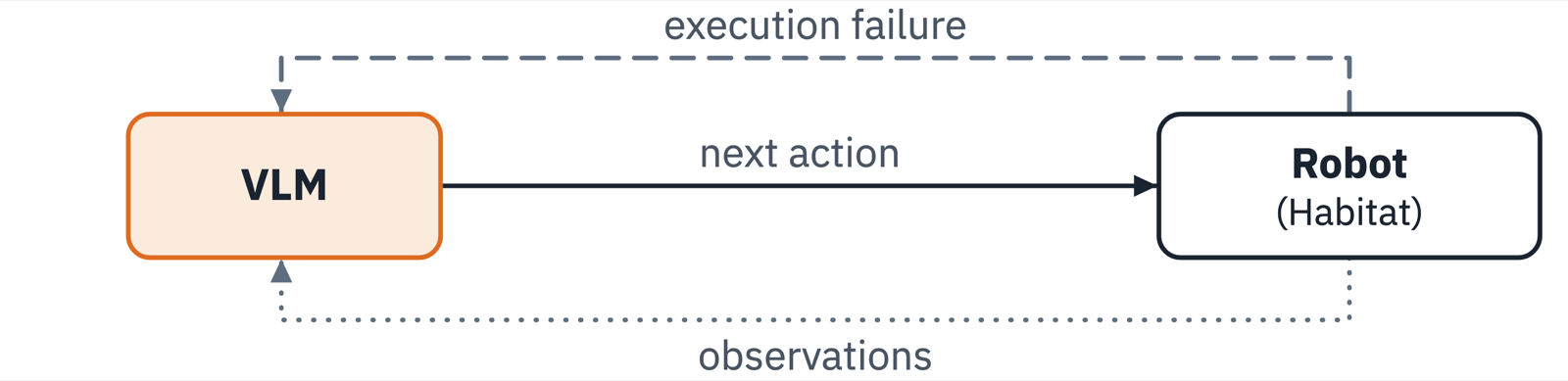}
    \caption{ReAct architecture.}
    \label{fig:diag-react}
\end{figure}

\begin{figure}[h]
    \centering
    \includegraphics[width=1\linewidth]{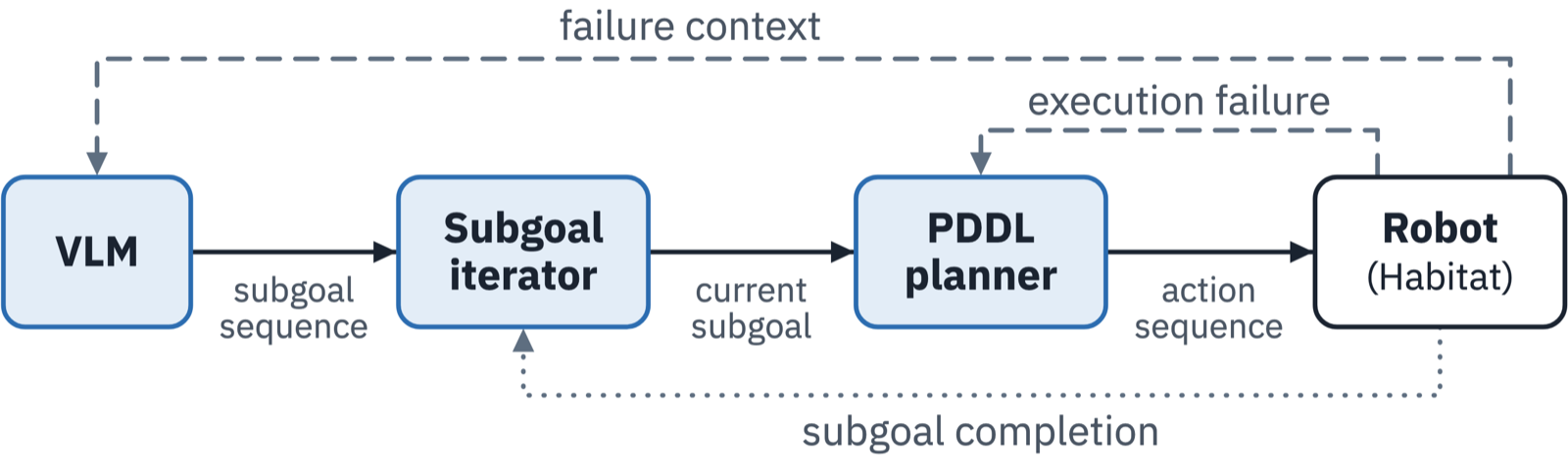}
    \caption{VLM-TAMP architecture.}
    \label{fig:diag-vlmtamp}
\end{figure}

\smallskip
\noindent \textbf{VLM-TAMP adaptation.}
We adapt the hierarchical planning approach of VLM-TAMP~\cite{vlmtamp-yang2024guiding} to Habitat as shown in Figure~\ref{fig:diag-vlmtamp}. Given task instructions, scene images, and the robot's current scene graph, a VLM generates intermediate subgoals expressed as grounded predicates. A symbolic planner then uses a hand-crafted PDDL domain~\cite{aeronautiques1998pddl} to generate executable skill sequences. Execution feedback triggers symbolic replanning or VLM-based subgoal revision when necessary. Our adaptation preserves this hierarchical architecture while using PARTNR's predefined skills for execution and incorporating the robot's initial memory and subsequent observations. We refer to this approach as \textbf{VLM-TAMP}.

\begin{figure}[h]
    \centering
    \includegraphics[width=1\linewidth]{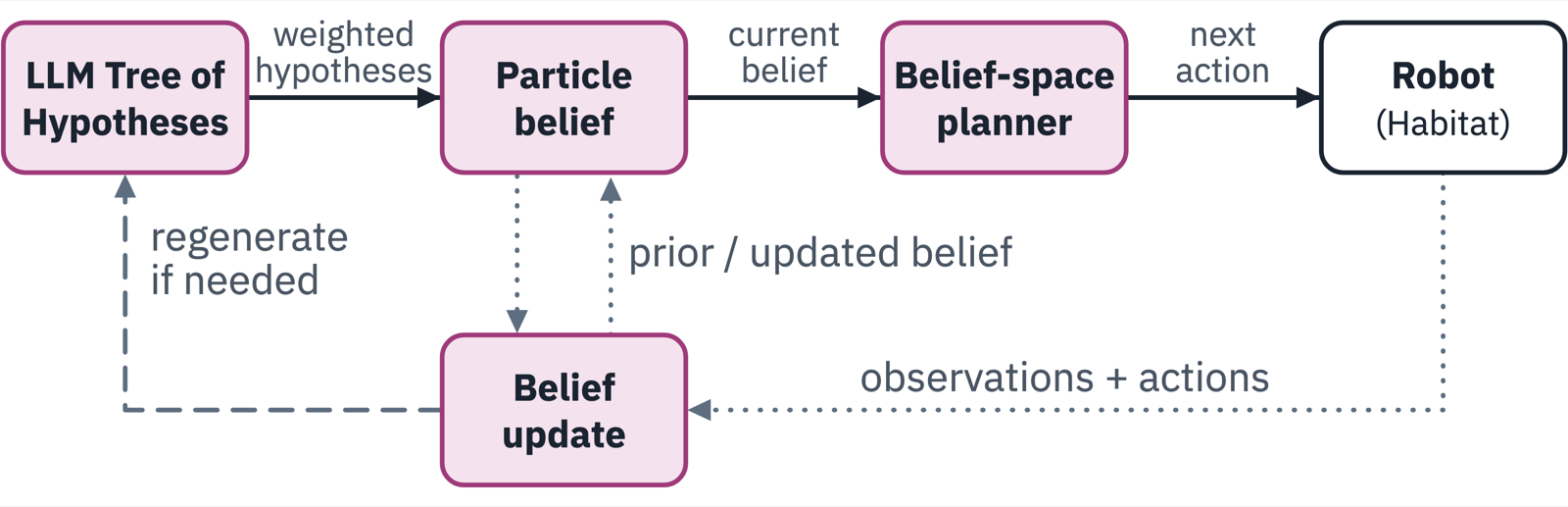}
    \caption{Tru-POMDP architecture.}
    \label{fig:diag-trupomdp}
\end{figure}

\smallskip
\noindent \textbf{Tru-POMDP adaptation.}
We adapt Tru-POMDP~\cite{tang2026trupomdptaskplanninguncertainty} to Habitat, as shown in Figure~\ref{fig:diag-trupomdp}. The approach maintains uncertainty over task goals and unobserved object locations using an LLM-based \textit{Tree of Hypotheses}, which constructs a weighted particle belief from textual scene observations, initial robot memory, and interaction history. Following each action, the belief is updated by predicting action effects, eliminating hypotheses inconsistent with observations, and generating new hypotheses when the remaining probability mass falls below a threshold. A DESPOT-style belief-tree search implemented in Python selects actions using a deterministic symbolic transition model and an adapted rollout policy. We extend the planning domain to support exploration and object-state changes. We refer to this adaptation as \textbf{Tru-POMDP}.

\smallskip
We evaluate all three approaches across seven household tasks, varying initial robot memory and environment configurations. Each applicable task--memory--environment combination is repeated five times per approach, yielding 1,350 episodes in total. All approaches use \textbf{GPT-5.6 Luna} with high reasoning effort as their underlying language or vision-language model, selected to balance planning accuracy, inference cost, and latency.

\section{Results and Discussion}
\label{sec:results}


We compare the three planning approaches in terms of overall task performance, sensitivity to initial memory and environmental variation, and recurring failure and inefficiency patterns. Together, these analyses characterize how different planning architectures handle uncertainty during long-horizon household tasks.

\subsection{Overall Task Performance}

\begin{figure}[H]
    \centering
    \includegraphics[width=0.95\linewidth]{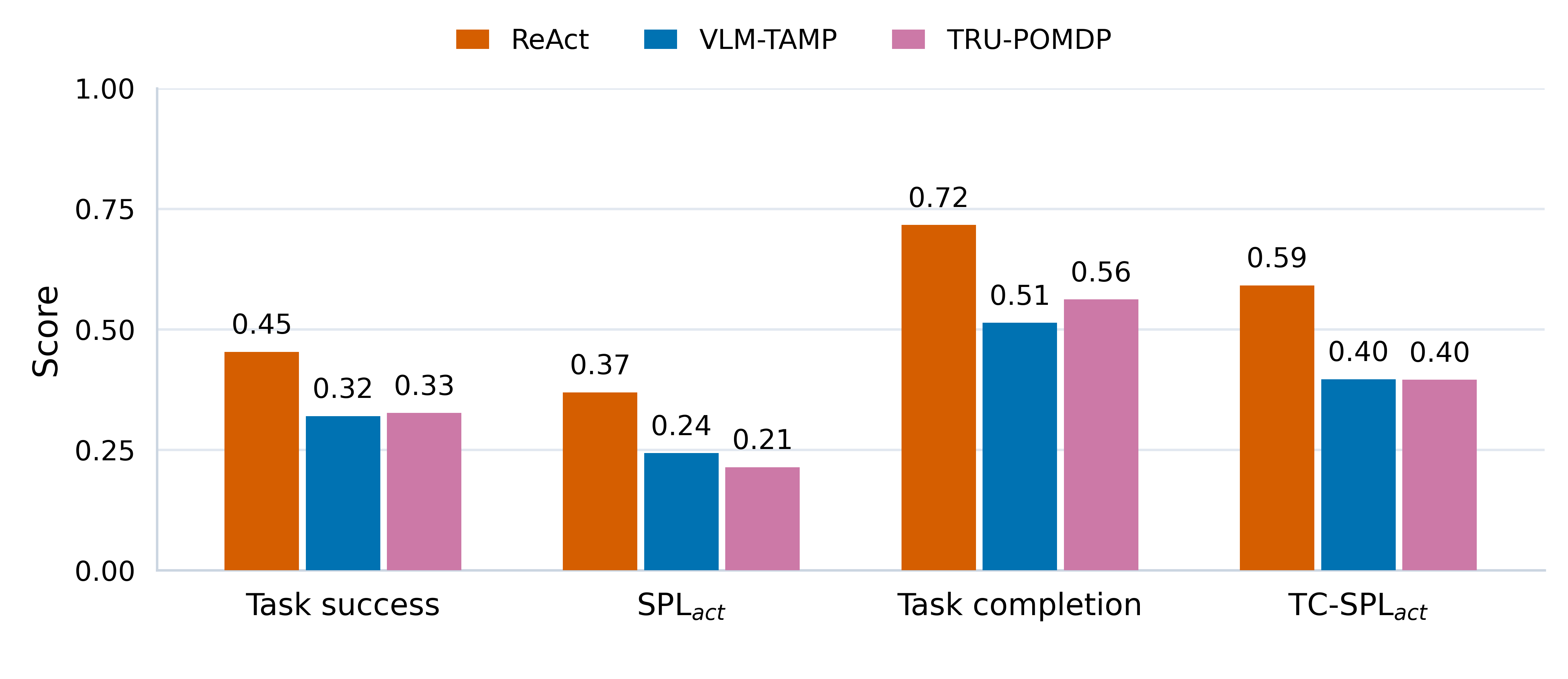}
    \caption{Overall task success, action-weighted SPL, task completion, and $\mathrm{TC\text{-}SPL}_{\mathrm{act}}$ across the three planning approaches.}
    \label{fig:overall_metrics}
\end{figure}

Figure~\ref{fig:overall_metrics} compares the three planners across task success, task completion, and action-weighted efficiency metrics. ReAct achieves the highest overall performance, with a task success rate of 45\% and mean task completion of 72\%, compared with 32\% and 51\% for VLM-TAMP, and 33\% and 56\% for Tru-POMDP. These results indicate that ReAct is more effective at completing long-horizon tasks than the two approaches incorporating explicit symbolic or belief-space planning.

However, incorporating action efficiency via the SPL metric results in lower performance scores across all three methods.
For ReAct, action-weighted SPL is 37\%, compared with a task success rate of 45\%, while $\mathrm{TC\text{-}SPL}_{\mathrm{act}}$ is 59\%, compared with 72\% task completion, showing a reduction.
Similar reductions occur for VLM-TAMP and Tru-POMDP, indicating that all three approaches execute additional skill attempts relative to human reference executions.

These findings highlight an important distinction between achieving task goals and executing them efficiently. Although ReAct achieves the highest overall performance, none of the evaluated approaches consistently produces efficient executions. We examine the sources of these inefficiencies in Section~\ref{sec:failure-analysis}.

\subsection{Effect of Initial Memory}


Table~\ref{tab:memory_tc_spl} reports $\mathrm{TC\text{-}SPL}_{\mathrm{act}}$ across the three initial memory conditions. The results reveal substantially different sensitivities to memory quality across planning approaches.

ReAct benefits strongly from accurate initial memory, achieving 0.71 compared with 0.55 under incomplete memory and 0.52 under outdated memory. This decline suggests that ReAct relies heavily on the correctness and completeness of its initial world representation, despite its ability to incorporate observations and action feedback during execution.

\begin{table}[H]
\centering
\caption{$\mathrm{TC\text{-}SPL}_{\mathit{act}}$ under different initial-memory conditions.}
\label{tab:memory_tc_spl}
\small
\setlength{\tabcolsep}{5pt}
\renewcommand{\arraystretch}{1.15}
\begin{tabular}{l|c|c|c}
\hline
Method & Accurate & Incomplete & Outdated \\
\hline
\textcolor{reactcolor}{ReAct}
& \textcolor{reactcolor}{0.71}
& \textcolor{reactcolor}{0.55}
& \textcolor{reactcolor}{0.52} \\
\hline
\textcolor{vlmcolor}{VLM-TAMP}
& \textcolor{vlmcolor}{0.37}
& \textcolor{vlmcolor}{0.39}
& \textcolor{vlmcolor}{0.43} \\
\hline
\textcolor{trucolor}{TRU-POMDP}
& \textcolor{trucolor}{0.39}
& \textcolor{trucolor}{0.32}
& \textcolor{trucolor}{0.47} \\
\hline
\end{tabular}
\end{table}

In contrast, VLM-TAMP and Tru-POMDP do not exhibit the same trend. VLM-TAMP achieves its highest score with outdated memory (0.43), compared with 0.37 under accurate memory. Similarly, Tru-POMDP achieves 0.47 with outdated memory, compared with 0.39 under accurate memory and 0.32 under incomplete memory. Thus, more accurate initial knowledge does not consistently translate into better task performance for these approaches.

One possible explanation is that discrepancies between memory and observations trigger additional exploration or replanning, potentially revealing information useful for completing the task. However, the aggregate results do not establish whether this mechanism accounts for the observed improvements. More generally, these findings suggest that planning performance depends not only on the quality of initial knowledge, but also on how a planner uses, verifies, and revises that knowledge during execution.



\subsection{Effect of Environmental Variations}

\begin{figure}[t]
    \centering
    \includegraphics[width=1\linewidth]{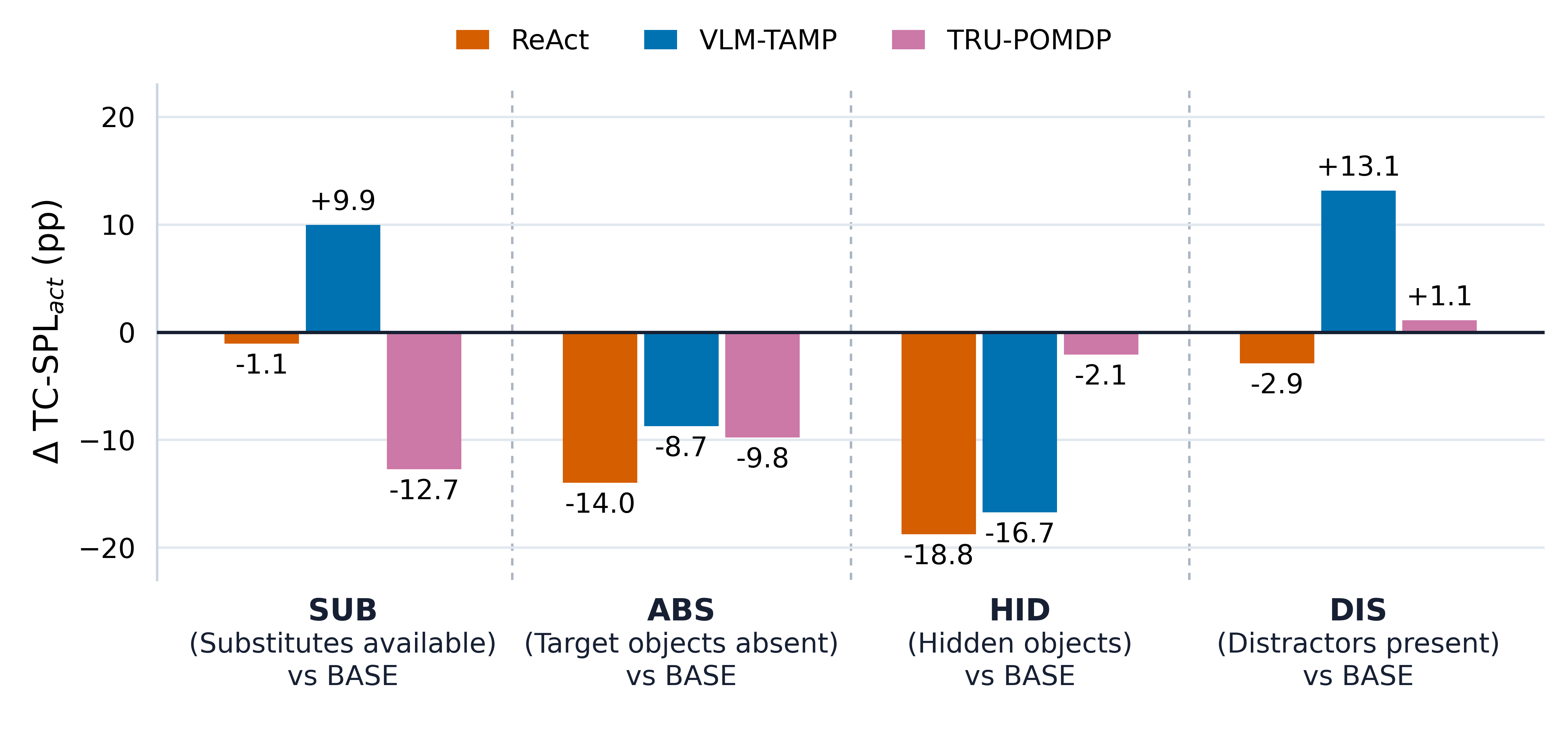}
    \caption{$\Delta \mathrm{TC\text{-}SPL}_{act}$ from BASE across environment perturbations, averaged over all three memory conditions and reported in percentage points.}
    \label{fig:memory_uncertainity_approaches}
\end{figure}

Figure~\ref{fig:memory_uncertainity_approaches} shows changes in $\mathrm{TC\text{-}SPL}_{\mathrm{act}}$ relative to the baseline environment (BASE), averaged across the three initial memory conditions. The results reveal two distinct patterns: environmental changes that consistently degrade performance and those whose effects depend strongly on the planning architecture.

\smallskip
\noindent \textbf{Absent and hidden objects.}
Removing target objects without providing suitable substitutes (ABS) reduces performance across all three planners, with the largest decrease observed for ReAct (14.0 percentage points). Similarly, hiding objects inside closed receptacles (HID) reduces performance for all methods, with ReAct experiencing the largest decline (18.8 percentage points). These configurations require planners to recognize that expected objects are unavailable or not directly accessible and to adjust their behavior accordingly.

Inspection of unsuccessful ReAct executions in ABS scenarios reveals instances of continued exploration and repeated visits to previously searched locations. Because no suitable object exists, these actions cannot resolve the missing-object condition and instead increase execution cost. This behavior illustrates a limitation of reactive replanning: adapting actions based on new observations does not necessarily provide an effective mechanism for determining when further exploration is unlikely to succeed.

\smallskip
\noindent \textbf{Substitutions and distractors.}
In contrast, substitutions (SUB) and distractors (DIS) produce markedly different outcomes across planners. With acceptable substitutes, VLM-TAMP improves by 9.9 percentage points relative to BASE, ReAct changes minimally (-1.1 points), and Tru-POMDP declines by 12.7 points. Similarly, distractors increase VLM-TAMP's score by 13.1 points, while producing smaller changes for ReAct (-2.9 points) and Tru-POMDP (+1.1 points).

\smallskip
These results indicate that environmental perturbations are not uniformly detrimental. Whereas hidden and absent targets consistently challenge all three approaches, substitutions and distractors interact differently with each planner. The improvements observed for VLM-TAMP are particularly noteworthy, although further trace-level analysis is needed to determine whether they reflect differences in object selection, subgoal generation, or other planning behaviors.

Overall, these findings demonstrate that robustness to environmental uncertainty cannot be characterized by a single performance measure. Different forms of uncertainty place different demands on exploration, object selection, and plan revision, with substantially different consequences across planning architectures.


\subsection{Failure and Inefficiency Analysis}
\label{sec:failure-analysis}

\begin{figure}[t]
    \centering
    \includegraphics[width=1\linewidth]{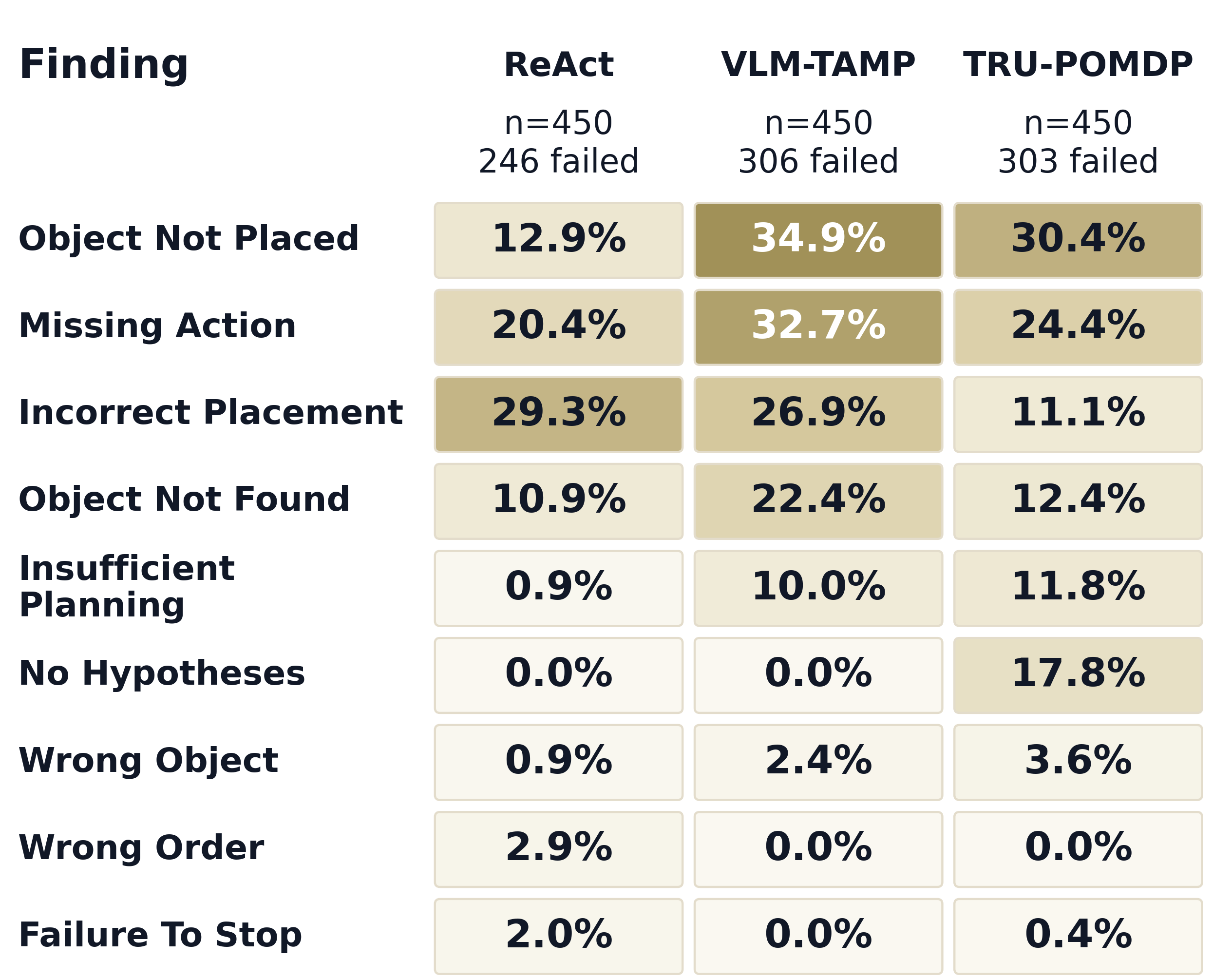}
    \caption{Failure-finding rates across planning approaches, reported as the percentage of all episodes exhibiting each failure; categories may overlap.}
    \label{fig:failure-findings}
\end{figure}

\begin{figure}[t]
    \centering
    \includegraphics[width=1\linewidth]{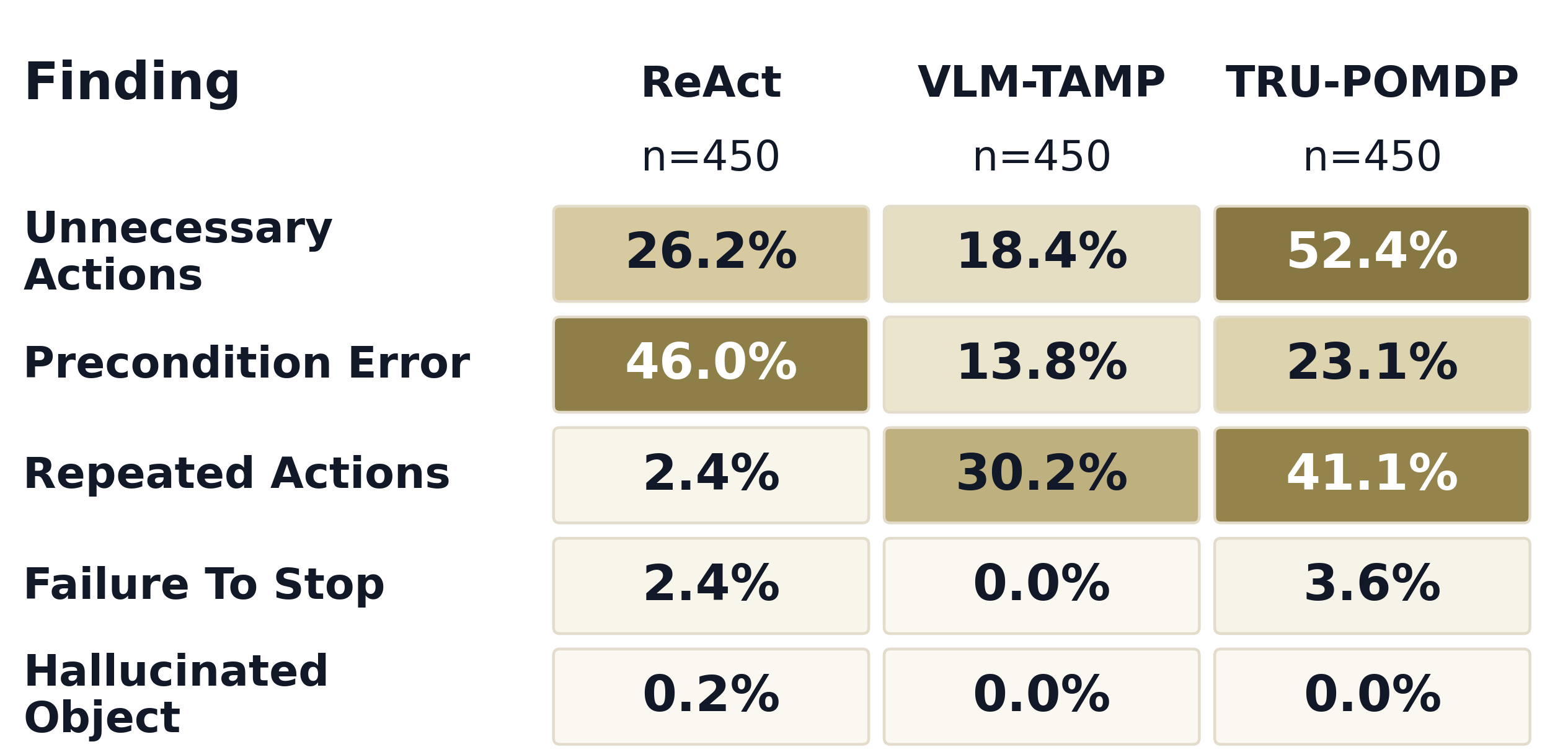}
    \caption{Inefficiency-finding rates across planning approaches, reported as the percentage of all episodes exhibiting each inefficiency; categories may overlap.}
    \label{fig:inefficiencies-findings}
\end{figure}

To better understand the performance differences, we analyze execution traces for recurring failure and inefficiency patterns. Figure~\ref{fig:failure-findings} and Figure~\ref{fig:inefficiencies-findings} report failure and inefficiency findings as percentages of all episodes. Because categories are not mutually exclusive, reported rates need not sum to 100\%.

\smallskip
\noindent \textbf{Task execution failures.}
The most prevalent failures involve objects not being placed at their required destinations, omitted actions, and incorrect placements. However, their relative frequencies differ substantially across planners.

VLM-TAMP exhibits the highest rate of object-not-placed findings (34.9\% of all episodes), compared with 12.9\% for ReAct and 30.4\% for Tru-POMDP. Inspection of VLM-TAMP traces reveals cases in which execution terminates after the generated skill sequence is exhausted, despite unmet task requirements. Although the system supports replanning when execution encounters difficulties, completing an executable subgoal sequence does not necessarily trigger verification of the overall task objective. This exposes a limitation of hierarchical planning when subgoal completion and task-level success are not consistently reconciled.

Missing-action findings are common across all three approaches (20.4--32.7\% of all episodes). These include omitted object-state changes, such as failing to turn off a lamp or clean a plate, even when other components of the task are completed. Such failures suggest that maintaining coverage of all task requirements remains challenging during long-horizon execution.

Placement errors are most frequent for ReAct (29.3\%), followed by VLM-TAMP (26.9\%) and Tru-POMDP (11.1\%). For example, in an inspected ReAct execution, the planner places a package on a side table rather than the requested center table, despite access to visual observations. This illustrates that visual grounding alone does not ensure correct destination selection. However, the lower placement-error rate for Tru-POMDP does not necessarily imply greater placement accuracy, as planners may differ in the number of placement actions attempted.

\smallskip
\noindent \textbf{Execution inefficiencies.}
Figure~\ref{fig:inefficiencies-findings} identifies three prominent inefficiency patterns: unnecessary actions, precondition errors, and repeated actions. These patterns reveal additional distinctions between the approaches.

Unnecessary actions occur in 18.4\% of all VLM-TAMP episodes, 26.2\% of ReAct episodes, and 52.4\% of Tru-POMDP episodes. Inspected traces include cases in which planners perform additional exploration despite already possessing accurate task-relevant information. This suggests that access to reliable memory does not necessarily prevent redundant information gathering.

Precondition errors are most frequent for ReAct (46.0\%), compared with 13.8\% for VLM-TAMP and 23.1\% for Tru-POMDP. For example, ReAct sometimes attempts to manipulate objects without first navigating to the object's location. VLM-TAMP's lower rate is consistent with the potential benefit of explicit symbolic precondition checking, although these results alone do not establish a causal relationship.

Repeated-action findings exhibit a contrasting pattern, occurring in 41.1\% of Tru-POMDP episodes and 30.2\% of VLM-TAMP episodes, but only 2.4\% of ReAct episodes. Thus, approaches with fewer precondition errors may nevertheless incur substantial inefficiency through repeated actions. In particular, Tru-POMDP's comparatively high rates of unnecessary and repeated actions suggest that explicit belief-space reasoning does not automatically translate into efficient exploration and execution.

\section{Summary and Takeaways}
Together, these analyses reveal distinct limitations across planning architectures:

\smallskip \noindent \textbf{ReAct} achieves the highest task performance but remains vulnerable to incorrect placements, precondition violations, and unproductive exploration when objects are unavailable. 

\smallskip \noindent \textbf{VLM-TAMP} reduces precondition errors but can terminate after completing generated subgoals without satisfying the full task. 

\smallskip \noindent \textbf{Tru-POMDP} explicitly represents uncertainty but exhibits frequent unnecessary and repeated actions.

\smallskip
These findings suggest that robust long-horizon planning requires more than reactive replanning, symbolic feasibility checking, or explicit uncertainty representation in isolation. Effective systems must also determine when additional information is needed, verify progress toward task-level goals, and recognize when continued execution or exploration is unlikely to be productive.

\section{Limitations and Conclusion}
\label{sec:conclusion}

Our evaluation focuses on high-level planning through a shared skill library. We do not separately characterize low-level execution failures, such as infeasible grasps, unreachable poses, or motion-control errors. The study is limited to seven simulated household tasks and our adaptations of three planning approaches. Furthermore, rule-based findings identify observable trace patterns rather than establish causal explanations.

The results show that planning performance depends on both initial robot memory and external environment variations, with different sensitivities across the three baselines. ReAct achieves the highest overall success and completion, but all approaches exhibit gaps between task performance and action efficiency.

As future work, we will develop planning systems that adaptively decide when to gather information, revise plans, or stop. We will evaluate these systems on our benchmark to assess whether they reduce failures and inefficiencies.



\bibliographystyle{IEEEtran}
\bibliography{references}

@article{coco-kim2026large,
  title={Large-language-model-guided state estimation for partially observable task and motion planning},
  author={Kim, Yoonwoo and Arora, Raghav and Mart{\'\i}n-Mart{\'\i}n, Roberto and Stone, Peter and Abbatematteo, Ben and Sung, Yoonchang},
  journal={arXiv preprint arXiv:2603.03704},
  year={2026}
}

@article{kaelbling2013integrated,
  title={Integrated task and motion planning in belief space},
  author={Kaelbling, Leslie Pack and Lozano-P{\'e}rez, Tom{\'a}s},
  journal={The International Journal of Robotics Research},
  volume={32},
  number={9-10},
  pages={1194--1227},
  year={2013},
  publisher={Sage Publications Sage UK: London, England}
}

@misc{garrett2020onlinereplanningbeliefspace,
      title={Online Replanning in Belief Space for Partially Observable Task and Motion Problems}, 
      author={Caelan Reed Garrett and Chris Paxton and Tomás Lozano-Pérez and Leslie Pack Kaelbling and Dieter Fox},
      year={2020},
      eprint={1911.04577},
      archivePrefix={arXiv},
      primaryClass={cs.RO},
      url={https://arxiv.org/abs/1911.04577}, 
}

@misc{clue-ravichandran2026activelyresolvingcontextualuncertainty,
      title={Actively Resolving Contextual Uncertainty for Underspecified Tasks in Natural Language}, 
      author={Zachary Ravichandran and Jonathan Diller and Fernando Cladera and Varun Murali and George J. Pappas and Vijay Kumar},
      year={2026},
      eprint={2609.30428},
      archivePrefix={arXiv},
      primaryClass={cs.RO},
      url={https://arxiv.org/abs/2609.30428}, 
}

@misc{CURE-yin2025reliablellmbasedrobotplanning,
      title={Towards Reliable LLM-based Robot Planning via Combined Uncertainty Estimation}, 
      author={Shiyuan Yin and Chenjia Bai and Zihao Zhang and Junwei Jin and Xinxin Zhang and Chi Zhang and Xuelong Li},
      year={2025},
      eprint={2510.08044},
      archivePrefix={arXiv},
      primaryClass={cs.RO},
      url={https://arxiv.org/abs/2510.08044}, 
}

@article{saycan-ahn2022can,
  title={Do as i can, not as i say: Grounding language in robotic affordances},
  author={Ahn, Michael and Brohan, Anthony and Brown, Noah and Chebotar, Yevgen and Cortes, Omar and David, Byron and Finn, Chelsea and Fu, Chuyuan and Gopalakrishnan, Keerthana and Hausman, Karol and others},
  journal={arXiv preprint arXiv:2204.01691},
  year={2022}
}

@misc{liu2023llmpempoweringlargelanguage,
      title={LLM+P: Empowering Large Language Models with Optimal Planning Proficiency}, 
      author={Bo Liu and Yuqian Jiang and Xiaohan Zhang and Qiang Liu and Shiqi Zhang and Joydeep Biswas and Peter Stone},
      year={2023},
      eprint={2304.11477},
      archivePrefix={arXiv},
      primaryClass={cs.AI},
      url={https://arxiv.org/abs/2304.11477}, 
}

@misc{ahn2025reliablecodeaspoliciesneurosymbolicframework,
      title={Towards Reliable Code-as-Policies: A Neuro-Symbolic Framework for Embodied Task Planning}, 
      author={Sanghyun Ahn and Wonje Choi and Junyong Lee and Jinwoo Park and Honguk Woo},
      year={2025},
      eprint={2510.21302},
      archivePrefix={arXiv},
      primaryClass={cs.AI},
      url={https://arxiv.org/abs/2510.21302}, 
}

@misc{singh2022progpromptgeneratingsituatedrobot,
      title={ProgPrompt: Generating Situated Robot Task Plans using Large Language Models}, 
      author={Ishika Singh and Valts Blukis and Arsalan Mousavian and Ankit Goyal and Danfei Xu and Jonathan Tremblay and Dieter Fox and Jesse Thomason and Animesh Garg},
      year={2022},
      eprint={2209.11302},
      archivePrefix={arXiv},
      primaryClass={cs.RO},
      url={https://arxiv.org/abs/2209.11302}, 
}

@article{vlmtamp-yang2024guiding,
  title={Guiding long-horizon task and motion planning with vision language models},
  author={Yang, Zhutian and Garrett, Caelan and Fox, Dieter and Lozano-P{\'e}rez, Tom{\'a}s and Kaelbling, Leslie Pack},
  journal={arXiv preprint arXiv:2410.02193},
  year={2024}
}

@article{garrett2021integrated,
  title={Integrated task and motion planning},
  author={Garrett, Caelan Reed and Chitnis, Rohan and Holladay, Rachel and Kim, Beomjoon and Silver, Tom and Kaelbling, Leslie Pack and Lozano-P{\'e}rez, Tom{\'a}s},
  journal={Annual review of control, robotics, and autonomous systems},
  volume={4},
  number={1},
  pages={265--293},
  year={2021},
  publisher={Annual Reviews}
}

@misc{puig2023habitat3,
      title  = {Habitat 3.0: A Co-Habitat for Humans, Avatars and Robots},
      author = {Xavi Puig and Eric Undersander and Andrew Szot and Mikael Dallaire Cote and Ruslan Partsey and Jimmy Yang and Ruta Desai and Alexander William Clegg and Michal Hlavac and Tiffany Min and Theo Gervet and Vladimír Vondru{\v{s}} and Vincent-Pierre Berges and John Turner and Oleksandr Maksymets and Zsolt Kira and Mrinal Kalakrishnan and Jitendra Malik and Devendra Singh Chaplot and Unnat Jain and Dhruv Batra and Akshara Rai and Roozbeh Mottaghi},
      year={2023},
      archivePrefix={arXiv},
}

@inproceedings{szot2021habitat,
  title     =     {Habitat 2.0: Training Home Assistants to Rearrange their Habitat},
  author    =     {Andrew Szot and Alex Clegg and Eric Undersander and Erik Wijmans and Yili Zhao and John Turner and Noah Maestre and Mustafa Mukadam and Devendra Chaplot and Oleksandr Maksymets and Aaron Gokaslan and Vladimir Vondrus and Sameer Dharur and Franziska Meier and Wojciech Galuba and Angel Chang and Zsolt Kira and Vladlen Koltun and Jitendra Malik and Manolis Savva and Dhruv Batra},
  booktitle =     {Advances in Neural Information Processing Systems (NeurIPS)},
  year      =     {2021}
}

@inproceedings{habitat19iccv,
  title     =     {Habitat: {A} {P}latform for {E}mbodied {AI} {R}esearch},
  author    =     {Manolis Savva and Abhishek Kadian and Oleksandr Maksymets and Yili Zhao and Erik Wijmans and Bhavana Jain and Julian Straub and Jia Liu and Vladlen Koltun and Jitendra Malik and Devi Parikh and Dhruv Batra},
  booktitle =     {Proceedings of the IEEE/CVF International Conference on Computer Vision (ICCV)},
  year      =     {2019}
}

@inproceedings{PARTNR,
  author = {Matthew Chang and Gunjan Chhablani and Alexander Clegg and Mikael Dallaire Cote and Ruta Desai and Michal Hlavac and Vladimir Karashchuk and Jacob Krantz and Roozbeh Mottaghi and Priyam Parashar and Siddharth Patki and Ishita Prasad and Xavier Puig and Akshara Rai and Ram Ramrakhya and Daniel Tran and Joanne Truong and John M. Turner and Eric Undersander and Tsung-Yen Yang},
  title = {PARTNR: A Benchmark for Planning and Reasoning in Embodied Multi-agent Tasks},
  booktitle = {International Conference on Learning Representations (ICLR)},
  note = {alphabetical author order},
  year = {2025}
}

@inproceedings{nayak2024LLaMAR,
    title={Long-Horizon Planning for Multi-Agent Robots in Partially Observable Environments},
    author={Nayak, Siddharth and Orozco, Adelmo Morrison and Ten Have, Marina and Zhang, Jackson and Thirumalai, Vittal and Chen, Darren and Kapoor, Aditya and Robinson, Eric and Gopalakrishnan, Karthik and Harrison, James and others},
    booktitle={The Thirty-eighth Annual Conference on Neural Information Processing Systems},
    year = {2024}
    }

@article{anderson2018evaluation,
  title={On evaluation of embodied navigation agents},
  author={Anderson, Peter and Chang, Angel and Chaplot, Devendra Singh and Dosovitskiy, Alexey and Gupta, Saurabh and Koltun, Vladlen and Kosecka, Jana and Malik, Jitendra and Mottaghi, Roozbeh and Savva, Manolis and others},
  journal={arXiv preprint arXiv:1807.06757},
  year={2018}
}

@misc{tang2026trupomdptaskplanninguncertainty,
      title={Tru-POMDP: Task Planning Under Uncertainty via Tree of Hypotheses and Open-Ended POMDPs}, 
      author={Wenjing Tang and Xinyu He and Yongxi Huang and Yunxiao Xiao and Cewu Lu and Panpan Cai},
      year={2026},
      eprint={2506.02860},
      archivePrefix={arXiv},
      primaryClass={cs.RO},
      url={https://arxiv.org/abs/2506.02860}, 
}

@misc{li2025embodiedagentinterfacebenchmarking,
      title={Embodied Agent Interface: Benchmarking LLMs for Embodied Decision Making}, 
      author={Manling Li and Shiyu Zhao and Qineng Wang and Kangrui Wang and Yu Zhou and Sanjana Srivastava and Cem Gokmen and Tony Lee and Li Erran Li and Ruohan Zhang and Weiyu Liu and Percy Liang and Li Fei-Fei and Jiayuan Mao and Jiajun Wu},
      year={2025},
      eprint={2410.07166},
      archivePrefix={arXiv},
      primaryClass={cs.CL},
      url={https://arxiv.org/abs/2410.07166}, 
}

@misc{yang2025embodiedbenchcomprehensivebenchmarkingmultimodal,
      title={EmbodiedBench: Comprehensive Benchmarking Multi-modal Large Language Models for Vision-Driven Embodied Agents}, 
      author={Rui Yang and Hanyang Chen and Junyu Zhang and Mark Zhao and Cheng Qian and Kangrui Wang and Qineng Wang and Teja Venkat Koripella and Marziyeh Movahedi and Manling Li and Heng Ji and Huan Zhang and Tong Zhang},
      year={2025},
      eprint={2502.09560},
      archivePrefix={arXiv},
      primaryClass={cs.AI},
      url={https://arxiv.org/abs/2502.09560}, 
}

@misc{yao2023reactsynergizingreasoningacting,
      title={ReAct: Synergizing Reasoning and Acting in Language Models}, 
      author={Shunyu Yao and Jeffrey Zhao and Dian Yu and Nan Du and Izhak Shafran and Karthik Narasimhan and Yuan Cao},
      year={2023},
      eprint={2210.03629},
      archivePrefix={arXiv},
      primaryClass={cs.CL},
      url={https://arxiv.org/abs/2210.03629}, 
}

@article{aeronautiques1998pddl,
  title={Pddl—the planning domain definition language},
  author={Aeronautiques, Constructions and Howe, Adele and Knoblock, Craig and McDermott, ISI Drew and Ram, Ashwin and Veloso, Manuela and Weld, Daniel and Sri, David Wilkins and Barrett, Anthony and Christianson, Dave and others},
  journal={Technical Report, Tech. Rep.},
  year={1998}
}

\newpage

\section{Appendix}
\label{sec:appendix}

\label{app:tasks-behavior}
\begingroup
\small
\subsection{Tasks}
These descriptions summarize BASE goals; other variants can change the required objects.

\begin{itemize}
\setlength{\itemsep}{0pt}
\setlength{\parsep}{0pt}

\item \textbf{T1 -- Water fetching.} Turn off all the lights in the bedroom and bring a full jug of water to either of the two living-room tables.

\item \textbf{T2 -- Package preparation.} Move both packages to the living-room center table and place them next to each other. Bring the scissors and place beside them to open the packages.

\item \textbf{T3 -- Book collection.} Collect the 2 green books \& place them on the living room side table and turn off the laptop as well.

\item \textbf{T4 -- Grocery organization.} Put all the 3 fruits in the fruit basket and the 2 drink containers on the kitchen counter.

\item \textbf{T5 -- Guest table preparation.} Get soap to clean both glasses, fill them, and place them on the living-room side table for our guests to drink. Get a white plate and place it next to them.

\item \textbf{T6 -- Meal delivery.} Heat up the bread, bring it to the desk in the bedroom the robot was initially at, bring a water bottle and a clean hand towel and place them next to the bread. Turn off the living-room lights.

\item \textbf{T7 -- Household cleanup.} Put away all toys from the living room into the closet's wardrobe, wash and put away the dirty dishes from sink inside, and turn off the living-room and kitchen lights.

\end{itemize}

\subsection{Failures}
A failure occurs when execution ends without satisfying the task's evaluation
criteria. The following labels describe behavior observed in failed runs.
\begin{itemize}
\setlength{\itemsep}{0pt}
\setlength{\parsep}{0pt}
\item \textbf{Placement error.} An object was placed in the wrong location.
\item \textbf{Object not placed.} The robot found an object but never put it where the task needs it: it never moved it, or was still holding it at the end.
\item \textbf{Wrong object.} The planner used a distractor instead of the actual task object. Approved substitutes are correct.
\item \textbf{Missing action.} A required action on an object was omitted or skipped, such as heating it or switching it off. An attempted action that failed to execute does not count.
\item \textbf{Wrong order.} A required step was done, but only after the step it had to come before.
\item \textbf{Insufficient planning.} The planner got stuck or continued unproductive execution instead of advancing.
\item \textbf{Object not found.} An object that a failed requirement needs was never found.
\item \textbf{No hypotheses.} TRU-POMDP only: the Tree of Hypotheses returned no usable goal guesses, so the planner stopped before taking any action.
\item \textbf{Failure to stop.} The planner kept trying to plan or search after evidence indicated that the subgoal should be abandoned.

\end{itemize}

\subsection{Inefficiencies}
An inefficiency is a behavior that wastes actions or execution effort without
useful progress toward the task. It can occur in both successful and failed runs.
\begin{itemize}
\setlength{\itemsep}{0pt}
\setlength{\parsep}{0pt}
\item \textbf{Hallucinated object.} Tried to use an object that does not exist. Using a stale object from outdated memory (OUT variants) does not count.
\item \textbf{Unnecessary actions.} Completed actions with no useful task, exploration, or recovery purpose.
\item \textbf{Precondition error.} Tried an action without valid prerequisites, such as picking an object before navigating within reach.
\item \textbf{Repeated actions.} The planner output the exact same action, or the exact same sequence of actions, several times in a row.
\item \textbf{Failure to stop.} Uses the definition above and is categorized as an
inefficiency when the run is successful.
\end{itemize}

\endgroup

\end{document}